\documentclass[journal=jacsat,manuscript=article]{achemso}

\usepackage{chemformula} 
\usepackage[T1]{fontenc} 
\usepackage{multirow}
\usepackage{amsmath,amssymb,amsfonts}
\usepackage{mathrsfs}
\usepackage{amsthm}
\usepackage{mathtools}
\usepackage{algorithm}
\usepackage{algorithmicx}
\usepackage{algpseudocode}
\usepackage{booktabs}
\usepackage{xr-hyper}
\usepackage{hyperref}
\usepackage{subcaption}
\usepackage{pifont} 
\usepackage{tikz}
\usepackage{threeparttable}
\usetikzlibrary{
    positioning,
    arrows.meta,
    shapes.geometric,
    fit,
    backgrounds,
    matrix,
    calc
}

\usepackage{listings}%
\usepackage{xcolor}
\usepackage{caption}                    
\usepackage{inconsolata}                
\definecolor{graybg}{HTML}{F6F7F8}      
\definecolor{framegray}{HTML}{BFC7CC}   
\definecolor{keywordcol}{HTML}{0B6FA4}  
\definecolor{stringcol}{HTML}{B45300}   
\definecolor{commentcol}{HTML}{6B7280}  
\definecolor{numbercol}{HTML}{9CA3AF}   

\hypersetup{
  colorlinks=true,
  linkcolor=blue,  
  citecolor=blue,  
  urlcolor=blue,   
  filecolor=blue,
  bookmarks=true
}

\lstdefinestyle{synrxnpy}{
  language=Python,
  basicstyle=\ttfamily\footnotesize,     
  keywordstyle=\bfseries\color{keywordcol},
  commentstyle=\itshape\color{commentcol},
  stringstyle=\color{stringcol},
  backgroundcolor=\color{graybg},
  frame=single,
  rulecolor=\color{framegray},
  framesep=6pt,                          
  xleftmargin=6pt,
  xrightmargin=6pt,
  aboveskip=6pt,
  belowskip=6pt,
  numbers=left,                          
  numberstyle=\scriptsize\color{numbercol},
  stepnumber=1,
  numbersep=8pt,
  showstringspaces=false,
  breaklines=true,
  breakatwhitespace=true,
  captionpos=b,                          
  columns=fullflexible,                  
  keepspaces=true,
  keepspaces=true,
  showtabs=false,
  tabsize=2
}

\lstdefinestyle{synrxnpy-nonum}{
  style=synrxnpy,
  numbers=none
}

\makeatletter
\def\moverlay{\mathpalette\mov@rlay}
\def\mov@rlay#1#2{\leavevmode\vtop{%
    \baselineskip\z@skip \lineskiplimit-\maxdimen
    \ialign{\hfil$\m@th#1##$\hfil\cr#2\crcr}}}
\newcommand{\charfusion}[3][\mathord]{
  #1{\ifx#1\mathop\vphantom{#2}\fi
    \mathpalette\mov@rlay{#2\cr#3}
  }
  \ifx#1\mathop\expandafter\displaylimits\fi}
\DeclareRobustCommand\bigop[1]{%
  \mathop{\vphantom{\sum}\mathpalette\bigop@{#1}}\slimits@
}
\newcommand{\bigop@}[2]{%
  \vcenter{%
    \sbox\z@{$#1\sum$}%
    \hbox{\resizebox{\ifx#1\displaystyle.9\fi\dimexpr\ht\z@+\dp\z@}{!}{$\m@th#2$}}%
  }%
}
\makeatother

\usepackage{cleveref} 
\usepackage{geometry} 
\usepackage{array} 
\usepackage{verbatim}
\usepackage{tabularx}
\usepackage{listings}
\usepackage{comment}
\usepackage{adjustbox}

\SectionNumbersOn

\theoremstyle{plain} 

\theoremstyle{definition} 

\usepackage{todonotes}

\newcommand{\synrxn}{\texttt{SynRXN}}
\newcommand{\taska}{\emph{reaction rebalancing}}
\newcommand{\taskb}{\emph{atom-to-atom mapping}}
\newcommand{\taskc}{\emph{reaction classification}}
\newcommand{\taskd}{\emph{reaction property prediction}}
\newcommand{\taske}{\emph{synthesis planning}}

\newcommand{\orcidicon}{\includegraphics[width=0.32cm]{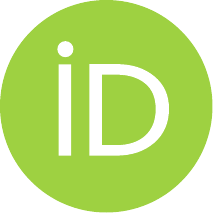}}

\foreach \x in {A, ..., Z}{%
\expandafter\xdef\csname orcid\x\endcsname{\noexpand\href{https://orcid.org/\csname orcidauthor\x\endcsname}{\noexpand\orcidicon}}
}

\usepackage{etoolbox} 

\let\oldminipage\minipage
\let\oldendminipage\endminipage
\renewenvironment{minipage}[1]{}{}
\makeatletter 
\g@addto@macro{\maketitle}{\let\minipage\oldminipage\let\endminipage\oldendminipage}
\makeatother

\author{Tieu-Long Phan\orcidD{}}
\email{tieu@bioinf.uni-leipzig.de}
\affiliation[Leipzig University]
{Bioinformatics Group, Department of Computer Science \& Interdisciplinary Center for Bioinformatics \& School for Embedded and Composite Artificial Intelligence (SECAI), Leipzig University, H{\"a}rtelstra{\ss}e 16–18, D-04107 Leipzig, Germany}
\alsoaffiliation[University of Southern Denmark]
{Department of Mathematics and Computer Science, University of Southern Denmark, DK-5230 Odense M, Denmark}

\author{Nhu-Ngoc Nguyen Song\orcidN{}}
\affiliation[UMP]
{School of Pharmacy, University of Medicine and Pharmacy at Ho Chi Minh City, Dinh Tien Hoang, Ho Chi Minh City, Vietnam}

\author{Peter F. Stadler\orcidB{}}
\affiliation[Leipzig University]
            {Bioinformatics Group, Department of Computer Science \& Interdisciplinary Center for Bioinformatics \& School for Embedded and Composite Artificial Intelligence (SECAI), Leipzig University, H{\"a}rtelstra{\ss}e 16–18, D-04107 Leipzig, Germany}

\alsoaffiliation[Max Planck Institute for Mathematics in the Sciences]
                {Max Planck Institute for Mathematics in the Sciences, Inselstra{\ss}e 22, D-04103, Leipzig, Germany}

\alsoaffiliation[University of Vienna]
                {Department of Theoretical Chemistry, University of Vienna, W{\"a}hringerstra{\ss}e 17, A-1090, Vienna, Austria}

\alsoaffiliation[Universidad National de Colombia]
                {Facultad de Ciencias, Universidad National de Colombia,Bogot{\'a}, Colombia}

\alsoaffiliation[University of Copenhagen]
{Center for non-coding RNA in Technology and Health, University of Copenhagen, Ridebanevej 9, DK-1870, Frederiksberg, Denmark}

\alsoaffiliation[Santa Fe Institute]
{Santa Fe Institute, 1399 Hyde Park Rd., Santa Fe, NM, 87501, USA}

\title[SynRXN]{\synrxn{}: \textbf{An Open Benchmark and Curated Dataset for Computational Reaction Modeling}}


\begin{document}

\maketitle






%
\begin{abstract}
We present \synrxn{}, a unified benchmarking framework and open-data resource for computer-aided synthesis planning (CASP). \synrxn{} decomposes end-to-end synthesis planning into five task families, covering reaction rebalancing, atom-to-atom mapping, reaction classification, reaction property prediction, and synthesis route design. Curated, provenance-tracked reaction corpora are assembled from heterogeneous public sources into a harmonized representation and packaged as versioned datasets for each task family, with explicit source metadata, licence tags, and machine-readable manifests that record checksums, and row counts. For every task, \synrxn{} provides transparent splitting functions that generate leakage-aware train, validation, and test partitions, together with standardized evaluation workflows and metric suites tailored to classification, regression, and structured prediction settings. For sensitive benchmarking, we combine public training and validation data with held-out gold-standard test sets, and contamination-prone tasks such as \taska{} and \taskb{} are distributed only as evaluation sets and are explicitly not intended for model training. Scripted build recipes enable bitwise-reproducible regeneration of all corpora across machines and over time, and the entire resource is released under permissive open licences to support reuse and extension. By removing dataset heterogeneity and packaging transparent, reusable evaluation scaffolding, \synrxn{} enables fair longitudinal comparison of CASP methods, supports rigorous ablations and stress tests along the full reaction-informatics pipeline, and lowers the barrier for practitioners who seek robust and comparable performance estimates for real-world synthesis planning workloads.
\end{abstract}

\section{Background and Summary}
\label{sec:background}

Computer-aided synthesis planning (CASP) assists chemists in designing feasible synthetic routes by combining mechanistic insight, experimental data, and algorithmic search. Recent advances in data-driven methods have improved single-step prediction, retrosynthesis, and multi-step route design, fueled by increasingly large, curated reaction corpora~\cite{coley2017, schwaller2018, segler2018, schwaller2020}. CASP is nevertheless a pipeline of distinct upstream and downstream components including extraction and curation, rebalancing, atom-atom mapping, classification, property prediction, and planning, each with its own assumptions and failure modes. While end-to-end evaluation is essential for system integration, rigorous isolated-task benchmarks deliver clear, comparable measures of progress for individual modules. Improvements at the task level propagate through the pipeline and produce more robust, reliable end-to-end workflows~\cite{genheden2022, maziarz2024}.

The pipeline begins with raw reaction data, which are often noisy or incomplete. Extractions from patents, electronic lab notebooks (ELNs), and the literature can omit solvents, counter-ions, stoichiometric reagents, or byproducts and can contain inconsistent stoichiometry or charge assignments~\cite{gimadiev2021reaction}. Such errors corrupt model inputs and bias learned representations~\cite{patel2009knowledge}. Automated \taska~methods such as \texttt{SynRBL}~\cite{phan2024} restore elemental and charge balance using rule-based and graph-based corrections, thereby producing cleaner inputs for subsequent stages. Because rebalancing is a corrective preprocessing step, standardized rebalancing test sets and diagnostics are necessary to quantify correction accuracy and to assess how residual inconsistencies may influence downstream tasks.

Once stoichiometry is addressed, \taskb~(AAM) reveals the microscopic changes that define each transformation. Accurate AAM is essential for identifying reaction centers, extracting mechanistic templates, and supervising models that reason about bond changes. Foundational studies have established robust automated mapping methods for complex reactions~\cite{jaworski2019}, and contemporary toolchains include transformer-based mappers such as \texttt{RXNMapper}~\cite{schwaller2021extraction}, graph-based mappers such as \texttt{GraphormerMapper}~\cite{nugmanov2022bidirectional} and \texttt{LocalMapper}~\cite{chen2024precise}, or heuristic mappers such as \texttt{RDTool}~\cite{rahman2016reaction}. Ensemble strategies that arbitrate among multiple mappers (for example, study of Li \emph{et al.}~\cite{lin2022atom} and \texttt{SynTemp}~\cite{phan2025syntemp}) increase coverage and flag low-confidence correspondences. Because mapping errors propagate into template extraction and mechanistic features, AAM benchmarking should rely on curated, held-out gold standards and report exact match accuracy.

When high-quality atom-to-atom mappings are available, template extraction and mechanistic clustering are straightforward. Because AAM is not yet universally reliable~\cite{jaworski2019} for many corpora, mapping-free classification remains practical and widely used, so benchmarks should evaluate both regimes and their sensitivity to mapping quality. \emph{Reaction classification} groups transformations by mechanism, functional-group change, or \emph{named-reaction} taxonomy, and supports search, curation, and downstream prediction. Methods range from engineered fingerprints and compact differential descriptors to learned embeddings. Engineered fingerprints enabled the first large-scale categorization efforts~\cite{schneider2015}. The compact, alignment-free descriptor \texttt{DRFP} encodes bond-change information via hashed circular substructure differences and remains competitive in small-data, interpretable settings~\cite{probst2022}. The learned representation \texttt{RXNFP} derives attention-based reaction embeddings from Molecular-Transformer sequence models~\cite{schwaller2019molecular}. Molecule-level cross-attention GNNs such as \texttt{SynCat} explicitly model inter-molecular context and reagent roles~\cite{van2025syncat}. Consequently, classification benchmarks require representative label taxonomies and transparent splitting procedures, together with repeated resampling to reveal robustness, calibration, and per-class behaviour across method families.

Beyond categorical labels, \emph{reaction property prediction} targets continuous and probabilistic quantities that guide experimental decisions. These include yields and physicochemical properties such as activation barriers and transition-state features. Graph- and sequence-based deep models have shown promise for yield and condition recommendations~\cite{schwaller2021yield, gao2018}, and representation-learning approaches (e.g., \emph{Condensed Graph of Reactions} or \emph{Imaginary Transition State} embeddings) extend to thermochemical and kinetic targets~\cite{heid2021}. Public barrier corpora such as \texttt{QMrxn/QMrxn20}~\cite{von2020thousands} and curated \texttt{SN2}/\texttt{E2} collections, and community benchmarking suites (e.g., \texttt{Chemprop} datasets~\cite{chemprop2023}) provide training and test splits for systematic barrier modeling and evaluation~\cite{spiekermann2022}. These studies show that machine learning predictors can approximate higher level quantum calculations when trained on high quality labels. However, they also reveal limited out-of-distribution transferability for transition-state properties, which motivates the use of QM-augmented descriptors and transition-state-based architectures~\cite{heid2021, spiekermann2022}.

The final stage is \emph{synthesis planning}, which composes one-step predictions into multi-step routes under feasibility, cost, and experimental constraints. Single-step models based on templates, sequence-to-sequence architectures, and graph neural networks have improved both forward prediction and retrosynthesis~\cite{coley2017, liu2017, schwaller2020}. Hybrid planners that combine learned policies~\cite{akhmetshin2024synplanner} with symbolic search produce efficient multi-step routes and expose explicit trade-offs between search budget and route quality~\cite{segler2018}. Evaluating planning algorithms is complex because route quality depends on budget, stopping criteria, route-cost models, and practical feasibility checks. Community efforts such as \texttt{PaRoutes}~\cite{genheden2022} and \texttt{Syntheseus}~\cite{maziarz2024} demonstrate the value of shared route corpora and standardized metrics, though broader harmonization across single- and multi-step settings remains necessary.

\begin{figure*}[htbp]
    \centering
    \includegraphics[width=1\textwidth]{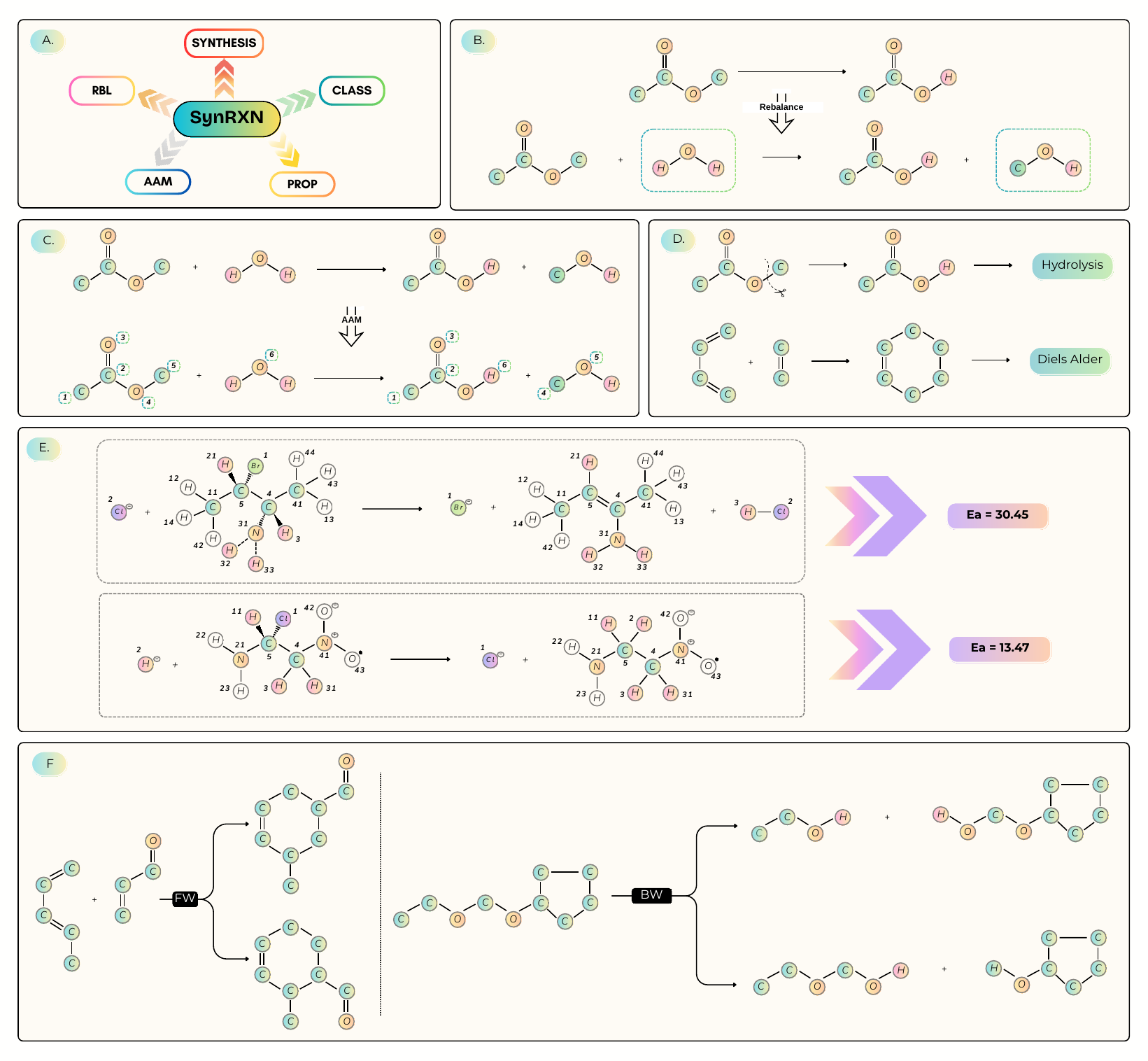}
     \caption{The \synrxn{} framework. (A) Benchmark suite overview. Individual tasks include (B) \taska, (C) \taskb, (D) \taskc, (E) \taskd, and (F) \taske. All task modules provide curated datasets, splitting functions, and evaluation metrics.}
     \label{fig:architecture}
\end{figure*}

These observations highlight a critical disparity: while the molecular machine learning community has flourished through standardized ecosystems like \emph{MoleculeNet}~\cite{wu2018} and the \emph{Therapeutics Data Commons}~\cite{huang2021}, reaction informatics remains fragmented. Current studies often rely on bespoke subsets, inconsistent preprocessing, and opaque splitting strategies, rendering cross-paper comparisons nearly impossible. Although broad aggregation efforts like the \emph{Open Reaction Database} (ORD)~\cite{kearnes2021open} solve the data \emph{access} problem, they do not provide the unified \emph{benchmarking} layer necessary to rigorously evaluate the full CASP pipeline. To bridge this gap, we introduce \synrxn{}, a unified, FAIR (Findable, Accessible, Interoperable, and Reusable)~\cite{wilkinson2016fair} benchmarking suite (see Figure~\ref{fig:architecture}). \synrxn{} goes beyond simple data hosting by enforcing strict reproducibility: it supplies deterministic splitting functions that, when paired with our versioned manifest files and RNG seeds, recreate identical train-test partitions without the overhead of massive index files. We further provide held-out gold standards for sensitive upstream tasks like rebalancing and atom-atom mapping (AAM). Available via \texttt{PyPI}\footnote{\url{https://pypi.org/project/synrxn/}} and \texttt{Zenodo}~\cite{phan2025zenodo}, \synrxn{} bundles standardized metrics, reference baselines, and CI-enabled regression checks. By codifying data hygiene and split transparency, \synrxn{} enables the first truly fair, head-to-head comparison of methods across the entire synthesis planning landscape.

\section{Methods}
\label{sec:methods}

The corpus is assembled from public repositories and community benchmarks, primarily the patent-extracted \texttt{USPTO}~\cite{lowe2012}, supplemented by \texttt{USPTO\_50K}~\cite{schneider2016} and larger USPTO compilations~\cite{jin2017wln}. We retain full provenance and licensing metadata, which are enumerated in a task-specific manifest. Preprocessing (standardization, normalization, optional rebalancing) and deterministic splitting functions (seed = 42) are implemented in the \synrxn{} package. Data, code, and manifests are released on \texttt{PyPI} and archived on \texttt{Zenodo}~\cite{phan2025zenodo}.

\subsection{Reaction rebalancing}
\label{subsec:rebalancing}

Chemical reaction records mined from patent literature frequently lack stoichiometric fidelity, often omitting necessary inorganic reagents, solvents, or byproducts. 
Restoring mass balance in these records is critical for downstream modeling (Figure~\ref{fig:architecture}B). 
To construct a chemically robust \taska{} benchmark, we derived targeted perturbation sets from the \texttt{USPTO\_50K} corpus~\cite{liu2017}, partitioning records into three specific modes of stoichiometric violation: \texttt{MNC} (Missing Non-Carbon), \texttt{MOS} (Missing One Side), and \texttt{MBS} (Missing Both Sides). 
Additionally, we curated a \texttt{Complex} set (1,892 examples) from manually validated \texttt{Golden} and Jaworski collections~\cite{lin2022atom,jaworski2019} to capture transformations involving significant skeletal rearrangements. 
The final compendium includes: \texttt{MNC} (33,147), \texttt{MOS} (12,781), \texttt{MBS} (491), and \texttt{Complex} (1,748) (Table~\ref{tab:dataset-stats}). 
Each subset was processed using \texttt{SynRBL}~\cite{phan2024}, a hybrid rule- and graph-based algorithm for reaction completion. 
To ensure high-quality ground truth, we retained only stoichiometric corrections resolved with a confidence of $\geq 90\%$.

\subsection{Atom-to-atom mapping}
\label{subsec:aam}

To rigorously benchmark \taskb~(Figure~\ref{fig:architecture}C), we stratified the reaction corpus into two distinct domains: \emph{small-molecule reactions} and \emph{biochemical transformations}.
The collection integrates high-fidelity reference sets, comprising the \texttt{Golden} dataset (1\,785)~\cite{lin2022atom,chen2024precise}, the manually curated Jaworski subset (491)~\cite{jaworski2019}, and a \texttt{USPTO\_3K} partition (3\,000) sampled from \texttt{USPTO\_50K}~\cite{chen2024precise}, totaling 5,276 reactions. 
The biochemical partition aggregates metabolic data from \texttt{Recon3D} (382)~\cite{litsa2018machine} and a validated \texttt{EColi} dataset (273)~\cite{beier2025computing}, yielding 655 reactions. 
Prior to mapping, all records were subjected to the \texttt{Standardize} and \texttt{CanonRSMI} modules of \texttt{SynKit}~\cite{phan2025synkit}. 
This pipeline enforces structural integrity by filtering anomalies (e.g., malformed SMILES strings, or valence/charge violations) and ensuring uniform canonicalization. 
Topological fidelity is evaluated using two criteria: graph isomorphism of \emph{Imaginary Transition State} (ITS) graphs~\cite{laffitte2023comparison,gonzalez2024partial,phan2025syntemp}, and exact-match accuracy of canonicalized reaction SMILES verified via \texttt{SynKit}. 
Dataset statistics are detailed in Table~\ref{tab:aam-stats}.

\begin{table}[htbp]
  \centering
  \caption{Rebalancing and atom-mapping benchmark datasets. Licenses refer to the redistributed, curated releases used in this work.}
  \label{tab:combined}
  \begin{subtable}[t]{0.45\textwidth}
    \centering
    \caption{Rebalancing task}
    \label{tab:dataset-stats}
    \footnotesize
    \begin{tabularx}{\linewidth}{@{} l r l >{\raggedleft\arraybackslash}X @{}}
      \toprule
      \textbf{Dataset}   & \textbf{Size}   & \textbf{License}   & \textbf{Reference} \\
      \midrule
      \texttt{MNC}      & 33\,147 & CC BY 4.0 &~\cite{liu2017} \\
      \texttt{MOS}      & 12\,781 & CC BY 4.0 &~\cite{liu2017} \\
      \texttt{MBS}      &    491  & CC BY 4.0 &~\cite{liu2017} \\
      \texttt{Complex}  &  1\,748 & CC BY 4.0 &~\cite{lin2022atom,jaworski2019} \\
      \midrule
    \end{tabularx}
  \end{subtable}
  \hfill
  \begin{subtable}[t]{0.5\textwidth}
    \centering
    \caption{Atom-mapping task}
    \label{tab:aam-stats}
    \footnotesize
    \setlength{\tabcolsep}{6pt}
    \begin{tabularx}{\linewidth}{@{} l l r l >{\raggedleft\arraybackslash}X @{}}
      \toprule
      \textbf{Dataset} & \textbf{Type} & \textbf{Size} & \textbf{License}   & \textbf{Reference} \\
      \midrule
      \texttt{Golden}    & Chem    & 1\,785 & CC BY 4.0 &~\cite{lin2022atom,chen2024precise} \\
      \texttt{NatComm}   & Chem    &   491 & CC BY 4.0 &~\cite{jaworski2019} \\
      \texttt{USPTO\_3K} & Chem    & 3\,000 & CC BY 4.0 &~\cite{chen2024precise} \\
      \texttt{Recon3D}   & Bio &   382 & CC BY 4.0 &~\cite{litsa2018machine} \\
      \texttt{EColi}     & Bio &   273 & CC BY 4.0 &~\cite{beier2025computing} \\
      \bottomrule
    \end{tabularx}
  \end{subtable}
\end{table}

\subsection{Reaction classification}
\label{subsec:class}

\emph{Reaction classification} requires mapping raw reaction inputs to predefined classes based on their structural or functional signature (Figure~\ref{fig:architecture}D). 
We assembled a benchmark suite spanning multiple levels of granularity. 
The \texttt{USPTO\_TPL} collection serves as a fine-grained standard (1,000 classes), annotated by deriving SMARTS templates from \texttt{RXNMapper} atom-maps~\cite{schwaller2021mapping,schwaller2021extraction}. 
For high-level categorization, we utilize the \texttt{Schneider} corpus (50 classes)~\cite{schneider2015}, which follows the hierarchical RSC reaction ontology. 
The \texttt{USPTO\_50K} dataset~\cite{lowe2012} is provided with two label sets: the legacy manual curation (10 classes)~\cite{liu2017} and a modern \textit{structural relabeling} via \texttt{SynTemp}~\cite{phan2025syntemp}. 
The latter enforces \textit{center-specific isomorphism} and extends the reaction core to controlled radii (R0--R2) to capture subtle mechanistic variations. 
Finally, biochemical diversity is addressed via \texttt{ECREACT}~\cite{zeng2025}, which provides enzyme-commission (EC) number hierarchies. 
To ensure rigorous evaluation, all corpora employ stratified splitting strategies that maintain \textit{label density} across all folds (see Table~\ref{tab:partition}).

\begin{table}[htbp]
  \centering
  \caption{Reaction-classification benchmarks. Splits are stratified by reaction class and generated deterministically. The \emph{Complete} column indicates whether reactions are fully specified or may be missing components.}
  \label{tab:partition}
  \footnotesize
  \setlength{\tabcolsep}{8pt}
  \begin{tabularx}{\linewidth}{@{} l c c c c l >{\raggedleft\arraybackslash}X @{}}
      \toprule
      \textbf{Dataset} & \textbf{Size} & \textbf{Split ratio} & \textbf{Classes} & \textbf{Complete} & \textbf{License} & \textbf{Reference} \\
      \midrule
      \texttt{Schneider\_U}     & 50\,000   & 9:1:40 & 50     & No  & CC BY 4.0 & \cite{schneider2015} \\
      \texttt{Schneider\_B}     & 50\,000   & 9:1:40 & 50     & Yes & CC BY 4.0 & \cite{schneider2015, van2025syncat} \\
      \texttt{USPTO\_TPL\_U}    & 445\,115  & 8:1:1  & 1\,000 & No  & CC BY 4.0 & \cite{schwaller2021mapping} \\
      \texttt{USPTO\_TPL\_B}    & 445\,115  & 8:1:1  & 1\,000 & Yes & CC BY 4.0 & \cite{schwaller2021mapping,van2025syncat} \\
      \texttt{USPTO\_50k\_U}    & 50\,016   & 8:1:1  & 10     & No  & CC BY 4.0 & \cite{liu2017} \\
      \texttt{USPTO\_50k\_B}    & 50\,016   & 8:1:1  & 10     & Yes & CC BY 4.0 & \cite{liu2017,van2025syncat} \\
      \texttt{SynTemp\_R0}      & 43\,441   & 8:1:1  & 143    & Yes & CC BY 4.0 & \cite{liu2017,phan2025syntemp} \\
      \texttt{SynTemp\_R1}      & 43\,441   & 8:1:1  & 356    & Yes & CC BY 4.0 & \cite{liu2017,phan2025syntemp} \\
      \texttt{SynTemp\_R2}      & 43\,441   & 8:1:1  & 680    & Yes & CC BY 4.0 & \cite{liu2017,phan2025syntemp} \\
      \texttt{ECREACT\_1st}      & 185\,734  & 8:1:1  & 7      & No  & CC BY 4.0 & \cite{zeng2025} \\
      \texttt{ECREACT\_2nd}      & 185\,734  & 8:1:1  & 63     & No  & CC BY 4.0 & \cite{zeng2025} \\
      \texttt{ECREACT\_3rd}      & 185\,734  & 8:1:1  & 175    & No  & CC BY 4.0 & \cite{zeng2025} \\
      \bottomrule
  \end{tabularx}
\end{table}

\subsection{Reaction Property prediction}
\label{subsec:property}

The \emph{Reaction Property Prediction} module (Figure~\ref{fig:architecture}E) targets the quantification of continuous chemical attributes.
We assembled a comprehensive benchmark suite by aggregating data from public repositories (including Zenodo) and literature. 
The suite encompasses \emph{ab initio kinetics} datasets (e.g., \texttt{B97XD3}, \texttt{LogRate}), specific mechanistic classes (\texttt{SNAr}, \texttt{SN2}, \texttt{E2}), and high-throughput experimental results (e.g., \texttt{RGD1}). 
A significant portion of the data was sourced from the \texttt{Heid} collection and related works~\cite{heid2021, heid_2023_10078142}, with additional datasets curated from the publications listed in Table~\ref{tab:reaction-property}. 
To ensure the integrity of the benchmark, all cleaning, standardization, and filtering operations are fully automated via \synrxn{} scripts. Datasets are provided in standardized formats containing either \texttt{rxn} (raw SMILES) or \texttt{aam} (atom-mapped SMILES) keys, mapped to specific property labels (e.g., ``eh'' for barrier height, ``dh'' for enthalpy, ``conversion'' for yield).

\begin{table}[htbp]
  \centering
  \begin{threeparttable}
  \caption{Reaction property datasets included in the \synrxn{} benchmark. The \emph{H} column indicates whether the dataset includes explicit hydrogen atoms. The \emph{Complete} column indicates whether reactions are fully specified (Yes) or may be missing components (No)}
  \label{tab:reaction-property}
  \footnotesize
  \setlength{\tabcolsep}{6pt}
  \begin{tabularx}{\linewidth}{@{} l r c c c c l >{\raggedright\arraybackslash}X @{}}
    \toprule
    \textbf{Dataset} & \textbf{Size} & \textbf{Split} & \textbf{AAM} & \textbf{H} & \textbf{Complete} & \textbf{License} & \textbf{Reference} \\
    \midrule
    \texttt{B97XD3}        & 16\,365  & 8:1:1 & Yes & Yes & No  & CC BY 4.0 & \cite{grambow2020reactants, grambow_2020_3715478} \\
    \texttt{SNAr}          &    503   & 8:1:1 & No  & No  & Yes & CC BY 3.0 & \cite{jorner2021machine} \\
    \texttt{E2SN2}         &  3\,625  & 8:1:1 & Yes & Yes & Yes & CC BY 4.0 & \cite{von2020thousands, heid2021} \\
    \texttt{Rad6Re}        & 31\,923  & 8:1:1 & Yes & Yes & Yes & CC BY 4.0 & \cite{stocker2020machine, heid2021} \\
    \texttt{LogRate}       &    778   & 8:1:1 & Yes & Yes & Yes & CC BY 4.0 & \cite{bhoorasingh2017automated, heid2021} \\
    \texttt{Phosphatase}   & 33\,354  & 8:1:1 & Yes & No  & Yes & CC BY 4.0\textsuperscript{a} & \cite{huang2015panoramic, heid2021} \\
    \texttt{E2}            &  1\,264  & 8:1:1 & Yes & Yes & Yes & CC BY 4.0 & \cite{heid_2023_10078142} \\
    \texttt{SN2}           &  2\,361  & 8:1:1 & Yes & Yes & Yes & CC BY 4.0 & \cite{heid_2023_10078142} \\
    \texttt{RDB7}          & 23\,852  & 8:1:1 & Yes & Yes & Yes & CC BY 4.0 & \cite{heid_2023_10078142} \\
    \texttt{CycloAdd}      &  5\,269  & 8:1:1 & Yes & Yes\textsuperscript{b} & Yes & CC BY 4.0 & \cite{heid_2023_10078142} \\
    \texttt{RGD1}          & 353\,984 & 8:1:1 & Yes & Yes & Yes & CC BY 4.0 & \cite{heid_2023_10078142} \\
    \bottomrule
  \end{tabularx}

  \begin{tablenotes}
    \footnotesize
    \item[a] Supplement to publication~\cite{heid2021}, released under the CC-BY~4.0 license.
    \item[b] Can be expanded to include explicit hydrogens (conversion available in our preprocessing scripts).
  \end{tablenotes}

  \end{threeparttable}
\end{table}

\subsection{Synthesis Planning}
\label{subsec:synplan}

The \emph{Synthesis Planning} module consolidates essential benchmarks for algorithmic route design. 
We rely on three established subsets of the \texttt{USPTO} patent literature: \texttt{USPTO\_50k}, the primary benchmark for template-based and template-free retrosynthesis~\cite{chen2021deep}; \texttt{USPTO\_MIT}, a high-volume corpus optimized for molecular transformer training~\cite{jin2017predicting}; and \texttt{USPTO\_500}, a specialized dataset targeting reagent and catalyst inference~\cite{lu2022unified}, which are summarized in Table~\ref{tab:synplan}. 
Crucially, we recommend standardized, deterministic splits to resolve prevalent issues with benchmark comparability. 
Our accompanying evaluation suite standardizes reporting protocols, focusing on conventional top-$k$ accuracy alongside structural similarity metrics.

\begin{table}[htbp]
  \centering
  \caption{Synthesis-planning corpora used in \synrxn.}
  \label{tab:synplan}
  \footnotesize
  \setlength{\tabcolsep}{6pt}
  \begin{tabularx}{\linewidth}{@{} l r l c l l >{\raggedleft\arraybackslash}X @{}}
    \toprule
    \textbf{Dataset} & \textbf{Size} & \textbf{Split} & \textbf{AAM} & \textbf{License} & \textbf{Task} & \textbf{Reference} \\
    \midrule
    \texttt{USPTO\_50k} & 50\,016   & 8:1:1      & Yes & CC BY 4.0 & forward / retrosynthesis & \cite{liu2017,chen2024precise} \\
    \texttt{USPTO\_MIT} & 479\,035  & 41:3:4     & Yes & CC BY 4.0 & forward / retrosynthesis & \cite{jin2017predicting} \\
    \texttt{USPTO\_500} & 143\,535  & 9:1:1.1    & No  & CC BY 4.0 & reagent prediction       & \cite{lu2022unified} \\
    \bottomrule
  \end{tabularx}
\end{table}

\section{Results}
\label{sec:results}

\subsection{Technical Validation}
\label{sec:validation}

\begin{figure*}[htbp]
  \centering
  \includegraphics[width=\textwidth]{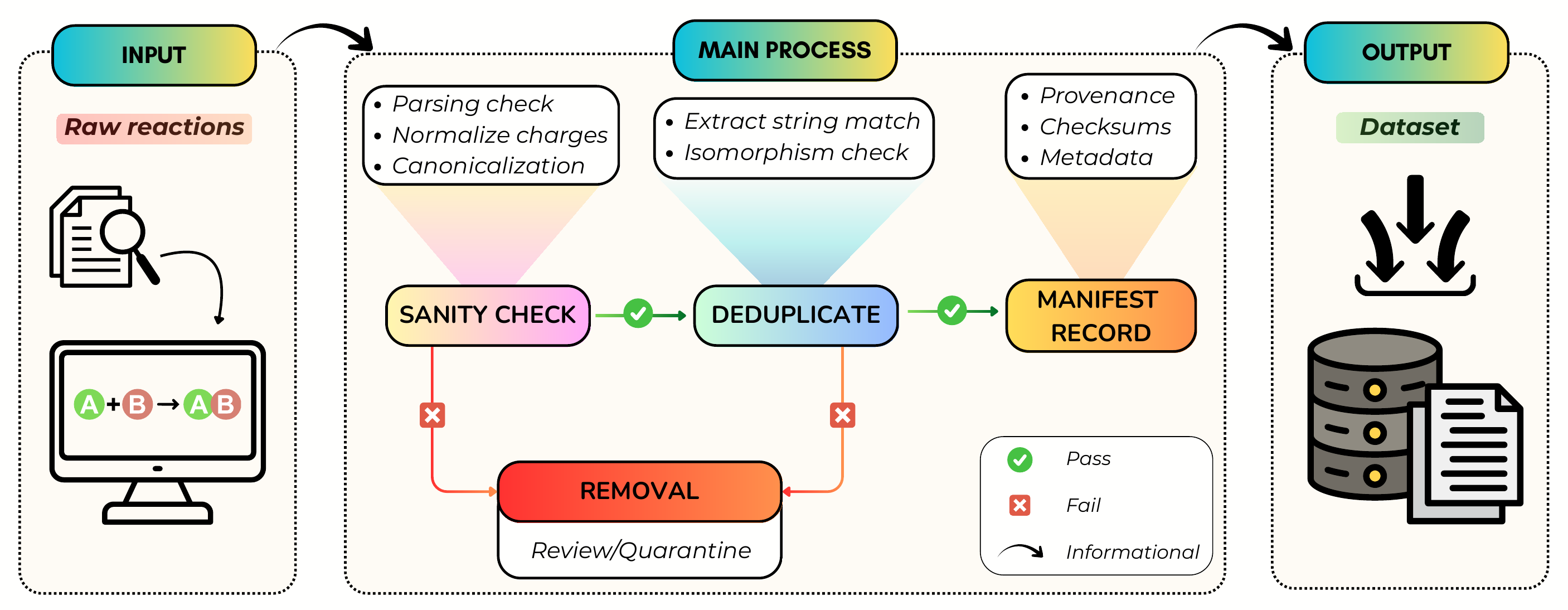}
  \caption{Technical validation workflow for the \synrxn{} benchmark.}
  \label{fig:tech}
\end{figure*}

Raw reaction records were retrieved from their original sources and ingested without manual curation or augmentation. Each entry passed through an automated, deterministic canonicalization and chemical-sanity pipeline implemented in \texttt{SynKit}~\cite{phan2025synkit}, which is illustrated in Figure~\ref{fig:tech}. The pipeline normalizes charges and valence states, standardizes aromaticity, and enforces a consistent SMILES canonicalization for all reactants, reagents, and products. We also perform automated duplicate detection during ingestion via (i) exact SMILES string matches and (ii) structure-level equivalence via isomorphism checks. Duplicate records identified by these checks are removed during ingestion; the dataset manifest records only the final counts so that the provenance and filtering outcome are reproducible. Records failing canonicalization or basic sanity checks (e.g.\ invalid SMILES, unparsable fields, impossible element/atom counts, or inconsistent valence) were excluded; no manual corrections were applied to excluded items. 

For the \taska~task, the \emph{test} split is curated to provide high-confidence ground truth: all candidate corrections proposed by \texttt{SynRBL} on the initial test pool were manually inspected, and only reactions that passed verification were retained, yielding \(100\%\) success rate and \(100\%\) exact-match accuracy that reflect manual selection of reliable labels rather than unconstrained automatic performance (see Section~\ref{si:rbl}). For atom-mapped subsets, we retain atom maps from the original sources and intentionally do not rebalance reactions, because current atom-mapping models are typically trained or pre-trained on raw, often incomplete data; mapped reactions are therefore only filtered by basic parsing and valence checks, such that invalid mapped SMILES or impossible valences are removed, while chemically plausible but stoichiometrically imbalanced reactions are kept, preserving the distributional characteristics of contemporary mapping corpora (baseline mapping accuracies for four tools across five held-out datasets are summarized in Table~\ref{tab:aam-baseline}). For datasets with numerical targets (e.g., yields, rates, or energies), we propagate the target values and their units exactly as provided by the original sources, each of which uses a single documented unit for the corresponding endpoint; any residual unit inconsistencies are thus inherited from the original data, and empirical baseline performance for these regression tasks, reported as MAE and MSE in Table~\ref{tab:baseline_prop}, confirms that the target distributions are learnable and free from obvious pathologies such as degenerate ranges or pervasive outliers.

To assess whether the benchmark tasks behave as intended, we provide reproducible baselines with fixed random seeds and fully specified preprocessing and training. Off-the-shelf \texttt{RandomForest} models on \texttt{DRFP}~\cite{probst2022} and \texttt{RXNFP}~\cite{schwaller2021mapping} embeddings perform robustly: reaction classification attains weighted \(\mathrm{F1}\) and multiclass MCC \(>0.9\) (Table~\ref{tab:class_stratify}), and property prediction reaches single-digit to low-tens MAE (Table~\ref{tab:baseline_prop}). Synthesis-planning metrics—coverage \(\mathcal{C}\), recognition rate (RR), \(F_{\beta}\), top-\(K\) \(\tau_K\), max-fragment overlap, and round-trip success—are defined in Section~\ref{si:synth}. Together, these baselines show the curated data are learnable yet non-trivial and provide stable reference points.

\subsection{Data}
\label{sec:datarecords}
\begin{figure*}[htbp]
  \centering
  \includegraphics[width=\textwidth]{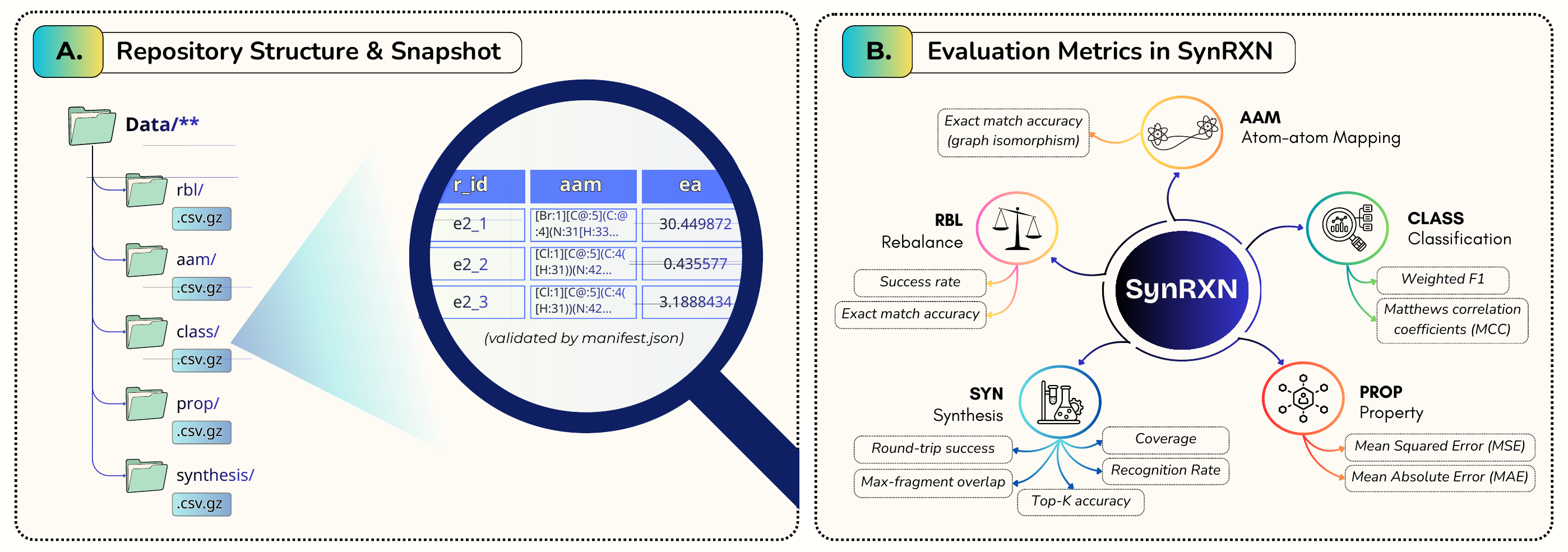}
  \caption{Overview of the \synrxn{} benchmark. (A) Data organization
  under the \texttt{Data/} root, showing task-specific subdirectories
  and main tabular records. (B) Evaluation metrics employed for the
  different \synrxn{} tasks.}
  \label{fig:data}
\end{figure*}

The \synrxn{} dataset is published as a machine-readable archive and source repository. Canonical releases are available from Zenodo (DOI: \texttt{10.5281/zenodo.17297258}) and mirrored on GitHub at \url{https://github.com/TieuLongPhan/SynRXN}. Each release
includes an authoritative \texttt{manifest.json} file that enumerates
all data files under the project root, recording for each file its
relative path, cryptographic checksum, row and column counts, column
names, a short human-readable description, and file-level license
information. This manifest serves as the primary source of provenance
and is used by the build and verification scripts to check the internal
consistency of each release. The top-level layout, file formats, and
evaluation metrics are summarized in Figure~\ref{fig:data}. All data
files in each release have an explicit license tag recorded in the
top-level \texttt{manifest.json} under the \texttt{license} field.

All data records reside in the top-level \texttt{Data/} directory, partitioned into task-specific subdirectories for the five benchmark tasks (Figure~\ref{fig:data}A): rebalancing (\texttt{rbl}), atom-atom mapping (\texttt{aam}), reaction classification (\texttt{class}), reaction property prediction (\texttt{prop}), and synthesis/retrosynthesis (\texttt{synthesis}). Each dataset is provided as a gzip-compressed CSV file (\texttt{.csv.gz}) in UTF-8 with a single header row and one reaction per decompressed line. Missing values are encoded as empty fields. Datasets include metadata and predefined train/validation/test splits supplied either as an in-table \texttt{split} column or as companion split files, and may contain task-specific columns such as \texttt{label}, \texttt{property}, mapping-completeness or hydrogen-explicitness flags, and external-source identifiers. Exact file-level metadata (paths, checksums, row/column counts, column names and license tags) are recorded in the top-level \texttt{manifest.json}.

Across all tasks, a small set of core columns is used consistently. The
\texttt{r\_id} column is a string that serves as a stable record
identifier (e.g.\ \texttt{uspto\_00001}), unique within each dataset.
The \texttt{rxn} column contains the canonical, unmapped reaction
SMILES, and \texttt{aam} holds the corresponding atom-mapped reaction
SMILES when available. Downstream task labels are stored in either a
\texttt{label} column (integer class codes for classification tasks) or
a \texttt{property} column (floating-point reaction properties such as
activation energies \texttt{ea}, enthalpies \texttt{dh}, or logarithmic
rate constants \texttt{lograte}). Some datasets include additional,
task-specific columns, for example flags indicating mapping completeness
or hydrogen-explicitness, or identifiers linking back to external
source corpora.

\subsubsection{Reaction rebalancing}
\label{si:rbl}
Rebalancing was performed with \texttt{SynRBL}. Each proposed output in the \emph{test} split was manually verified, and only reactions that passed verification were retained in the published test set. Consequently, on the released test set \texttt{SynRBL} attains \(100\%\) \emph{success rate} and \(100\%\) \emph{exact match} accuracy; this reflects curation/selection rather than unconstrained automatic performance. Candidate and ground-truth reactions are canonicalized with \texttt{SynKit}. We report: (i) \emph{success rate}, the fraction of inputs whose rebalanced output is element-wise stoichiometrically balanced; and \emph{exact match}, the fraction whose \emph{canonical} reaction SMILES exactly equals the curated ground truth.

\subsubsection{Atom-to-atom mapping}
\label{si:aam}
We benchmarked four atom-mappers across five strictly held-out datasets (see Table~\ref{tab:aam-baseline}). For each reaction we apply the same deterministic standardization pipeline to both the predicted and curated SMILES using \texttt{synkit}, and we convert the standardized reactions to ITS graphs for comparison. Mapping quality is assessed by a graph isomorphism test using \texttt{synkit.Chem.Reaction.aam\_validator} which verifies that predicted atom correspondences reproduce the curated ITS graph. 

\begin{table}[htbp]
  \centering
  \caption{Atom-mapping accuracy (\%) across five datasets.}
  \label{tab:aam-baseline}
  \setlength{\tabcolsep}{10pt}%
  \renewcommand{\arraystretch}{1.15}%
  \begin{tabularx}{\textwidth}{@{} l *{5}{>{\centering\arraybackslash}X} @{}}
    \toprule
      & \texttt{EColi} & \texttt{Recon3D} & \texttt{USPTO\_3K} & \texttt{Golden} & \texttt{NatComm} \\
    \midrule
    \texttt{RXNMapper 0.4.1}      & 72.53 & 48.69 & 93.53 & 87.43 & 87.58 \\
    \texttt{Graphormer}\textsuperscript{*} & 42.12 & 34.82 & 95.10 & 89.59 & 92.87 \\
    \texttt{LocalMapper 0.1.5}  & 69.96 & 50.79 & 97.77 & 89.08 & 92.67 \\
    \texttt{RDTool 2.4.1}                  & 78.02 & 54.97 & 90.87 & 82.54 & 84.11 \\
    \bottomrule
  \end{tabularx}
  \vspace{4pt}
  \begin{minipage}{\textwidth}
    \footnotesize\raggedright
    \textsuperscript{*}\,\texttt{Graphormer} built with \texttt{Cython 1.7.8}.
  \end{minipage}
\end{table}

\subsubsection{Reaction classification}
\label{si:class}
For classification we compared \texttt{RXNFP} and \texttt{DRFP} embeddings used as fixed input features to a \texttt{RandomForest} classifier, which is from \texttt{scikit-learn}~\cite{kramer2016scikit}, and evaluated performance using a repeated stratified cross-validation protocol (5 repeats of 5-fold CV, i.e.\ $5{\times}5$ k\/-fold) and follow the statistical testing procedure of Ash \emph{et al.}~\cite{ash2025}. Stratified folds preserve class proportions and mitigate variance due to class imbalance. Primary evaluation metrics are the weighted  $F_1$ score and the multiclass Matthews correlation coefficient (MCC). The benchmarking results for the stratified split are reported in Table~\ref{tab:class_stratify}.
\begin{equation}\label{eq:f1_weighted_complete}
  \mathrm{F1_{weighted}}
  = \sum_{i=1}^{K} \frac{n_i}{N}\,\mathrm{F1}_i,
  \qquad
  \mathrm{F1}_i = \frac{2\,\mathrm{Precision}_i\,\mathrm{Recall}_i}{\mathrm{Precision}_i+\mathrm{Recall}_i},
  \qquad
  N=\sum_{j=1}^{K} n_j,
  \end{equation}
  
  \begin{equation}\label{eq:prec_recall_tp_fp_fn}
  \mathrm{Precision}_i=\frac{\mathrm{TP}_i}{\mathrm{TP}_i+\mathrm{FP}_i},
  \qquad
  \mathrm{Recall}_i=\frac{\mathrm{TP}_i}{\mathrm{TP}_i+\mathrm{FN}_i},
  \end{equation}
  
  \begin{equation}\label{eq:tp_fp_fn_confmat}
  \mathrm{TP}_i = C_{ii},\qquad
  \mathrm{FP}_i = \sum_{r=1}^{K} C_{ri} - C_{ii},\qquad
  \mathrm{FN}_i = \sum_{c=1}^{K} C_{ic} - C_{ii},
  \end{equation}
  
  \begin{equation}\label{eq:mcc_multi_complete}
  \mathrm{MCC_{multi}} =
  \frac{c\,s - \sum_{k=1}^{K} p_k t_k}
  {\sqrt{\bigl(s^{2}-\sum_{k=1}^{K} p_k^{2}\bigr)\bigl(s^{2}-\sum_{k=1}^{K} t_k^{2}\bigr)}},
  \end{equation}
  
  \begin{equation}\label{eq:mcc_defs_complete}
  c=\sum_{k=1}^{K} C_{kk},\qquad
  s=\sum_{i=1}^{K}\sum_{j=1}^{K} C_{ij},\qquad
  p_k=\sum_{i=1}^{K} C_{ik},\qquad
  t_k=\sum_{j=1}^{K} C_{kj},
  \end{equation}
  
  \noindent where \(C\) is the \(K\times K\) confusion matrix with entry \(C_{ij}\) equal to the count of true class \(i\) predicted as class \(j\), \(n_i\) is the number of true instances of class \(i\) (so \(n_i=\sum_{j}C_{ij}\)), and \(\mathrm{TP}_i,\mathrm{FP}_i,\mathrm{FN}_i\) are true/false positives/negatives for class \(i\) as defined above.
  
  \begin{table}[htbp]
    \centering
    \caption{Reaction classification on stratified splits (5${\times}$5 stratified CV, seed = 42) comparing \texttt{DRFP} and \texttt{RXNFP} with a \texttt{RandomForest}. Values are mean~$\pm$~std; better mean in \textbf{bold}. Significance: NS ($p>0.05$), * ($p<0.05$), ** ($p<0.01$), *** ($p<0.001$), **** ($p<0.0001$).}
    \label{tab:class_stratify}
    \scriptsize
    \setlength{\tabcolsep}{4pt}%
    \renewcommand{\arraystretch}{1.06}%
    \newcolumntype{L}[1]{>{\raggedright\arraybackslash}p{#1}}%
    \newcolumntype{C}[1]{>{\centering\arraybackslash}p{#1}}%
    \begin{tabularx}{\linewidth}{@{} L{0.13\linewidth} C{0.03\linewidth}
                                  C{0.15\linewidth} C{0.15\linewidth} C{0.04\linewidth}
                                  C{0.15\linewidth} C{0.15\linewidth} C{0.04\linewidth} @{}}
    \toprule
    Dataset & Level
      & \multicolumn{3}{c}{\(\mathrm{F1_{weighted}}\uparrow\)} & \multicolumn{3}{c}{\(\mathrm{MCC}\uparrow\)} \\
    \cmidrule(lr){3-5} \cmidrule(lr){6-8}
      & & \texttt{DRFP} & \texttt{RXNFP} & \(p\) & \texttt{DRFP} & \texttt{RXNFP} & \(p\) \\
    \midrule
    \texttt{Schneider\_U}   & -- & \textbf{0.968 $\pm$ 0.002} & 0.962 $\pm$ 0.002 & **** & \textbf{0.968 $\pm$ 0.002} & 0.961 $\pm$ 0.002 & **** \\
    \texttt{Schneider\_B}   & -- & \textbf{0.953 $\pm$ 0.002} & 0.936 $\pm$ 0.002 & **** & \textbf{0.952 $\pm$ 0.002} & 0.935 $\pm$ 0.002 & **** \\
    \texttt{USPTO\_TPL\_U}   & -- & \textbf{0.968 $\pm$ 0.002} & 0.962 $\pm$ 0.002 & **** & \textbf{0.968 $\pm$ 0.002} & 0.961 $\pm$ 0.002 & **** \\
    \texttt{USPTO\_TPL\_B}   & -- & \textbf{0.953 $\pm$ 0.002} & 0.936 $\pm$ 0.002 & **** & \textbf{0.952 $\pm$ 0.002} & 0.935 $\pm$ 0.002 & **** \\
    \texttt{USPTO\_50k\_U}   & -- & 0.953 $\pm$ 0.002 & \textbf{0.958 $\pm$ 0.002} & **** & 0.943 $\pm$ 0.003 & \textbf{0.949 $\pm$ 0.002} & **** \\
    \texttt{USPTO\_50k\_B}   & -- & \textbf{0.966 $\pm$ 0.002} & 0.952 $\pm$ 0.002 & **** & \textbf{0.958 $\pm$ 0.002} & 0.941 $\pm$ 0.002 & **** \\
    \texttt{SynTemp}         & 0  & \textbf{0.952 $\pm$ 0.001} & 0.920 $\pm$ 0.002 & **** & \textbf{0.954 $\pm$ 0.001} & 0.927 $\pm$ 0.002 & **** \\
    \texttt{SynTemp}         & 1  & \textbf{0.940 $\pm$ 0.002} & 0.897 $\pm$ 0.002 & **** & \textbf{0.943 $\pm$ 0.002} & 0.903 $\pm$ 0.002 & **** \\
    \texttt{SynTemp}         & 2  & \textbf{0.913 $\pm$ 0.003} & 0.737 $\pm$ 0.004 & **** & \textbf{0.907 $\pm$ 0.003} & 0.714 $\pm$ 0.005 & **** \\
    \texttt{ECREACT}         & 1  & \textbf{0.977 $\pm$ 0.001} & 0.905 $\pm$ 0.001 & **** & \textbf{0.966 $\pm$ 0.001} & 0.862 $\pm$ 0.002 & **** \\
    \texttt{ECREACT}         & 2  & \textbf{0.964 $\pm$ 0.001} & 0.857 $\pm$ 0.002 & **** & \textbf{0.961 $\pm$ 0.001} & 0.846 $\pm$ 0.002 & **** \\
    \texttt{ECREACT}         & 3  & \textbf{0.949 $\pm$ 0.001} & 0.840 $\pm$ 0.001 & **** & \textbf{0.947 $\pm$ 0.001} & 0.835 $\pm$ 0.001 & **** \\
    \bottomrule
    \end{tabularx}
  \end{table}
  
\subsubsection{Reaction property prediction}
\label{si:prop}
We benchmark property prediction using \texttt{DRFP} and \texttt{RXNFP} embeddings with \texttt{scikit-learn}~\cite{kramer2016scikit}'s \texttt{RandomForestRegressor}. Evaluation uses repeated cross-validation (5 repeats of 5 folds, i.e., 25 runs), and results are reported as mean~$\pm$~std. For the large \texttt{RGD1} dataset (353,984 examples) we use a single train/validation/test split; to estimate variance we repeat this experiment with five different random seeds and follow the statistical testing procedure of Ash \emph{et al.}~\cite{ash2025}. Test-set MAE and MSE are reported in Table~\ref{tab:baseline_prop}.

\begin{equation}\label{eq:mae}
\mathrm{MAE} \;=\; \frac{1}{N}\sum_{i=1}^N \bigl|y_i - \hat{y}_i\bigr|
\end{equation}

\begin{equation}\label{eq:mse}
\mathrm{MSE} \;=\; \frac{1}{N}\sum_{i=1}^N \bigl(y_i - \hat{y}_i\bigr)^2
\end{equation}

\begin{table}[ht]
  \centering
  \caption{Reaction property prediction on random splits comparing \texttt{DRFP} and \texttt{RXNFP} with a \texttt{RandomForest}. Values are mean~$\pm$~std; better mean in \textbf{bold}. Significance: NS ($p>0.05$), * ($p<0.05$), ** ($p<0.01$), *** ($p<0.001$), **** ($p<0.0001$).}
  \label{tab:baseline_prop}
  \scriptsize
  \setlength{\tabcolsep}{6pt}
  \renewcommand{\arraystretch}{1.05}
  \newcolumntype{P}[1]{>{\centering\arraybackslash}p{#1}}
  \begin{tabularx}{\linewidth}{@{} l l P{0.14\linewidth} P{0.14\linewidth} P{0.05\linewidth} P{0.14\linewidth} P{0.14\linewidth} P{0.05\linewidth} @{}}
  \toprule
  Dataset & Prop
    & \multicolumn{3}{c}{MAE $\downarrow$} & \multicolumn{3}{c}{MSE $\downarrow$} \\
  \cmidrule(lr){3-5} \cmidrule(lr){6-8}
    & & \texttt{DRFP} & \texttt{RXNFP} & \(p\) & \texttt{DRFP} & \texttt{RXNFP} & \(p\) \\
  \midrule
  \texttt{B97XD3}    & dh       & 19.838 $\pm$ 0.262 & \textbf{19.323 $\pm$ 0.214} & **** & 649.814 $\pm$ 21.305 & \textbf{599.521 $\pm$ 15.667} & **** \\
  \texttt{B97XD3}    & ea       & \textbf{14.617 $\pm$ 0.268} & 15.324 $\pm$ 0.239 & **** & \textbf{376.803 $\pm$ 13.839} & 396.723 $\pm$ 12.360 & **** \\
  \texttt{CycloAdd}  & act      & \textbf{5.853 $\pm$ 0.157}  & 6.115 $\pm$ 0.157  & **** & \textbf{57.696 $\pm$ 4.071} & 63.569 $\pm$ 4.261 & **** \\
  \texttt{CycloAdd}  & r        & \textbf{11.790 $\pm$ 0.306} & 12.081 $\pm$ 0.312 & **** & \textbf{227.691 $\pm$ 12.297} & 236.306 $\pm$ 12.085 & ** \\
  \texttt{E2}        & ea       & \textbf{3.247 $\pm$ 0.206}  & 7.377 $\pm$ 0.354  & **** & \textbf{20.067 $\pm$ 3.174} & 91.161 $\pm$ 8.376 & **** \\
  \texttt{E2SN2}     & ea       & \textbf{4.150 $\pm$ 0.126}  & 7.116 $\pm$ 0.133  & **** & \textbf{30.667 $\pm$ 2.074} & 81.454 $\pm$ 2.857 & **** \\
  \texttt{LogRate}   & lograte  & \textbf{1.054 $\pm$ 0.068}  & 1.077 $\pm$ 0.059  & NS  & \textbf{1.970 $\pm$ 0.284} & 2.149 $\pm$ 0.355 & ** \\
  \texttt{Phosphatase} & Conversion & \textbf{0.098 $\pm$ 0.001} & 0.099 $\pm$ 0.001 & **** & 0.019 $\pm$ 0.000 & 0.019 $\pm$ 0.000 & **** \\
  \texttt{Rad6Re}    & dh       & 1.126 $\pm$ 0.019 & \textbf{0.908 $\pm$ 0.013} & **** & 2.585 $\pm$ 0.083 & \textbf{1.612 $\pm$ 0.052} & **** \\
  \texttt{RDB7}      & ea       & 30.136 $\pm$ 0.210 & \textbf{18.812 $\pm$ 0.240} & **** & 1362.068 $\pm$ 16.817 & \textbf{579.031 $\pm$ 15.282} & **** \\
  \texttt{RGD1}      & ea       & 16.704 $\pm$ 0.074 & \textbf{15.953 $\pm$ 0.032} & NS  & 495.386 $\pm$ 3.867 & \textbf{453.876 $\pm$ 2.628} & NS \\
  \texttt{SN2}       & ea       & \textbf{4.433 $\pm$ 0.161}  & 6.940 $\pm$ 0.234  & **** & \textbf{34.664 $\pm$ 2.426} & 75.566 $\pm$ 4.393 & **** \\
  \texttt{SNAr}      & ea       & \textbf{1.402 $\pm$ 0.158}  & 1.447 $\pm$ 0.139  & NS  & \textbf{4.348 $\pm$ 1.496} & 4.355 $\pm$ 1.032 & NS \\
  \bottomrule
  \end{tabularx}
\end{table}

\subsubsection{Synthesis planning}
\label{si:synth}
We provide benchmarking datasets, lightweight data loaders, and split/evaluation
scripts for synthesis planning, but do not reimplement full-scale synthesis
planners due to computational and dependency constraints. For published
benchmark results, see Han \emph{et al.}~\cite{han2024retrosynthesis}
(\texttt{USPTO\_50k}, \texttt{USPTO\_MIT}) and Lu \& Zhang~\cite{lu2022unified}
(\texttt{USPTO\_500}). \synrxn{} including data loaders and extended evaluation metrics are formally described here to facilitate reproducibility and future method development.

Let \(N\) be the number of test problems. For problem \(i\), the canonicalized
ground-truth route is \(R_i\), and a planner returns \(M_i\) ranked candidate
routes \(R'_{i,1},\dots,R'_{i,M_i}\). All equality tests use the same
deterministic canonicalization and, where applicable, ITS-graph isomorphism.
Denote the indicator function by \(\mathbb{I}(\cdot)\). Define
\(\mathrm{frag}(\cdot)\) to return a route's canonical fragment set,
\(\mathrm{reactants}(\cdot)\) to extract reactants,
\(\mathrm{target}(\cdot)\) to extract the canonical target product(s), and
\(\mathrm{forward}(\cdot)\) to map reactants to a predicted product via an
exact reaction application or forward model.

\begin{equation}
\label{eq:coverage_eq}
\mathcal{C}
= \frac{1}{N}\sum_{i=1}^{N}\mathbb{I}\bigl(R_i = R'_{i,1}\bigr)
\end{equation}
Coverage \(\mathcal{C}\) is a route-level \emph{recall} metric: for each
problem \(i\), we check whether the planner returns the canonical ground-truth
route anywhere in its candidate list, and average this indicator across all
problems. A value of \(\mathcal{C}=1\) means that every ground-truth route is
hit at least once.

\begin{equation}
\label{eq:rr_eq}
\mathrm{RR}
= \frac{1}{N}\sum_{i=1}^{N}\frac{1}{M_i}\sum_{j=1}^{M_i}
   \mathbb{I}\bigl(R_i = R'_{i,j}\bigr)
\end{equation}
Recognition rate \(\mathrm{RR}\) is the per-problem average fraction of returned
candidates that exactly match the ground truth. It behaves as a
\emph{precision} metric at the route level: given the list of routes a
planner proposes for problem \(i\), \(\mathrm{RR}\) measures what proportion of
those proposals are the canonical ground truth, and then averages this
per-candidate correctness across problems, independent of ranking.

\begin{equation}
\label{eq:fbeta_eq}
F_{\beta}
= \frac{(1+\beta^2)\,\bigl(\mathrm{RR}\cdot\mathcal{C}\bigr)}
       {\beta^2\cdot\mathrm{RR} + \mathcal{C}}
\end{equation}
The \(F_{\beta}\)-score combines \(\mathrm{RR}\) (precision-like) and
\(\mathcal{C}\) (recall-like) in the usual \(F_\beta\) form; \(\beta=1\) gives
\(F_1\). Choosing \(\beta>1\) emphasizes recall (\(\mathcal{C}\)), while
\(\beta<1\) emphasizes precision (\(\mathrm{RR}\)).

\begin{equation}
\label{eq:topk_eq}
\tau_K
= \frac{1}{N}\sum_{i=1}^{N}
  \mathbb{I}\!\Bigl(R_i \in \{R'_{i,1},\dots,R'_{i,K}\}\Bigr)
\end{equation}
Top-\(K\) accuracy \(\tau_K\) is the fraction of problems where the ground truth
appears among the first \(K\) predictions (typical \(K\): 1, 3, 5). It assesses
whether correct routes are returned within a short candidate list even if not
top-ranked.

\begin{equation}
\label{eq:maxfrag_i_eq}
\mathrm{max\_frag}_i
= \max_{1\le j\le M_i}
    \frac{|\mathrm{frag}(R_i)\cap\mathrm{frag}(R'_{i,j})|}
         {|\mathrm{frag}(R_i)|}
\end{equation}
\begin{equation}
\label{eq:maxfrag_eq}
\mathrm{max\_frag}
= \frac{1}{N}\sum_{i=1}^{N}\mathrm{max\_frag}_i
\end{equation}
Max-fragment overlap \(\mathrm{max\_frag}\) measures partial correctness by
selecting, for each problem, the candidate that recovers the largest fraction of
ground-truth fragments (reagents/intermediates or another chosen fragment
definition) and averaging across problems. This captures partial agreement when
exact route matching is too strict.
\begin{equation}
\label{eq:round_cand_eq}
\mathrm{round}_{i,j}
= \mathbb{I}\!\bigl(
    \mathrm{forward}(\mathrm{reactants}(R'_{i,j}))
    = \mathrm{target}(R_i)
  \bigr)
\end{equation}
\begin{equation}
\label{eq:round_prob_eq}
\mathrm{round}_i
= \max_{1\le j\le M_i}\mathrm{round}_{i,j},
\qquad
\rho = \frac{1}{N}\sum_{i=1}^{N}\mathrm{round}_i
\end{equation}
Round-trip success \(\rho\) checks chemical consistency: for each candidate, we
apply the forward check \(\mathrm{forward}\) to the candidate's reactants. If
any candidate reproduces the target product(s), the problem counts as
successful. \(\rho\) detects chemically consistent precursors even when
string-level equality fails.

\clearpage
\subsection{Usage Notes}
\label{sec:usage}

The \synrxn{} loader supports three sources for programmatic access: \textbf{Zenodo} (stable, citable archive and recommended for publications), \textbf{GitHub release tag} (release artifacts) and \textbf{GitHub commit} (exact snapshot; may be unstable unless archived). Note that Zenodo queries can occasionally be delayed; enable GitHub fallback when immediate access is required.

\begin{lstlisting}[style=synrxnpy,caption={Load \texttt{SynRXN} data},label={lst:dl-example},captionpos=t]
from synrxn.data import DataLoader

dl = DataLoader(
    task="prop",          # "rbl" | "aam" | "class" | "prop" | "synthesis"
    source="zenodo",      # "zenodo" | "github" | "commit"
    version="0.0.8",      # "zenodo" version | "github" tag | "commit" id
    gh_enable=True,       # allow GitHub fallback when Zenodo is slow
)
df = dl.load(name = "b97xd3")
\end{lstlisting}

We recommend using \textbf{Zenodo} for publication workflows (set \texttt{source="zenodo"} and provide \texttt{version}). Use \texttt{source="github"} with a release tag for release-centric workflows. For exact reproducibility use \texttt{source="commit"} and provide the full \texttt{commit\_id}; if those results are published, archive the snapshot (create a GitHub release or deposit on Zenodo) so it is citable. For splitting strategies, we recommend the code snippet below for transparent splitting and reproducibility:

\begin{lstlisting}[style=synrxnpy-nonum,caption={Deterministic repeated K-fold splitting (reproducible)},label={lst:split-example},captionpos=t]
from synrxn.split.repeated_kfold import RepeatedKFoldsSplitter
# Load the dataset
df = dl.load("b97xd3")   

# Construct a reproducible repeated K-Fold splitter
splitter = RepeatedKFoldsSplitter(
    n_splits=5,         # folds per repeat
    n_repeats=5,        # number of repeats
    ratio=(8, 1, 1),    # train : val : test proportions
    shuffle=True,       # shuffle before splitting
    random_state=42,    # deterministic seed for reproducibility
)

# Compute splits (optionally stratify for classification tasks)
splitter.prepare_splits(df, stratify=None)

# Retrieve a specific repeat/fold as DataFrames
train_df, val_df, test_df = splitter.get_split(repeat=0, fold=0, as_frame=True)

\end{lstlisting}

For full end-to-end reproducibility and to verify the published archives, you
can rebuild datasets locally using the CLI:

\begin{lstlisting}[style=synrxnpy-nonum,caption={Rebuild datasets using the \synrxn{}}, label={lst:data-gen},captionpos=t]
# from repository root (editable install recommended)
python -m synrxn --build
# no-save (if supported) to preview actions without writing files
python -m synrxn build --property --no-save
\end{lstlisting}

\section{Conclusion}
\label{sec:conclusion}

In this work, we introduced \synrxn{}, a provenance-aware and versioned benchmarking framework designed to standardize CASP. By decomposing CASP into five core task families---ranging from reaction rebalancing to full retrosynthetic analysis---we provide the community with a harmonized, machine-readable resource that prioritizes data integrity. Our framework addresses long-standing challenges in the field through leakage-aware splitters, reproducible build recipes, and explicit metadata tracking. Technical validation via the \texttt{SynKit} pipeline ensures that our datasets are not only consistent but also transparent regarding their limitations. We believe \synrxn{} establishes a fair and verifiable baseline for future CASP methods. To support immediate community adoption, the framework can be installed via \texttt{pip install synrxn}. The complete dataset, manifests, and codebase are publicly available via \texttt{Zenodo} (\url{https://doi.org/10.5281/zenodo.17297258}) and \texttt{GitHub} (\url{https://github.com/TieuLongPhan/SynRXN}).

\clearpage

\section*{Conflicts of interest}
There are no conflicts to declare.

\section*{Disclaimer}
Views and opinions expressed are however those of the author(s) only and do not necessarily reflect those of the European Union. Neither the European Union nor the granting authority can be held responsible for them.

\section*{Code availability}
The \texttt{SynRXN} source code is available on \texttt{GitHub}: \url{https://github.com/TieuLongPhan/SynRXN}. Archived releases are available on \texttt{Zenodo} (DOI: 10.5281/zenodo.17672847; archived release: \url{https://zenodo.org/record/17672847}). Comprehensive software documentation is hosted at \url{https://synrxn.readthedocs.io/en/latest/}.

\section*{Data availability}
All data supporting this study are included in the project repository under the \texttt{Data} directory in \texttt{GitHub} \url{https://github.com/TieuLongPhan/SynRXN} and archived releases on \texttt{Zenodo} \url{https://zenodo.org/record/17672847}. 

\section*{Funding}
This project has received funding from the European Unions Horizon Europe
Doctoral Network programme under the Marie-Sk{\l}odowska-Curie grant
agreement No~101072930 (TACsy -- Training Alliance for Computational
systems chemistry).

\begin{acknowledgement}

We also thank Ngoc-Vi Nguyen Tran for commenting on the manuscript. 

\end{acknowledgement}

\section*{Authors' contributions}
Contributions are reported according to the CRediT taxonomy.
T.L.P. conceptualized the study, curated and analyzed the data, developed the methods and software, validated the results, and wrote and revised the manuscript. N.N.N.S. designed figures, conducted benchmarking, and reported baseline results. P.F.S. secured funding and resources, supervised the project, and drafted and reviewed the manuscript.







\bibliography{achemso-demo}

\end{document}